\documentclass[final]{cvpr}

\usepackage{times}
\usepackage{epsfig}
\usepackage{graphicx}
\usepackage{amsmath}
\usepackage{amssymb}

\usepackage{ifthen}
\usepackage{color}
\usepackage{bigstrut}
\usepackage{multirow}
\usepackage{multicol}
\usepackage{mathtools}
\usepackage[normalem]{ulem}
\usepackage{algorithm}
\usepackage{algpseudocode}

\newcommand{\myfootnote}[2]{{%
  \let\thempfn\relax
  \footnotetext[0]{$^#1$\emph{#2}}
}}

\DeclarePairedDelimiter{\norm}{\lVert}{\rVert}
\usepackage[pagebackref=true,breaklinks=true,colorlinks,bookmarks=false,citecolor=blue,linkcolor=blue]{hyperref}

\definecolor{frenchblue}{rgb}{0.0, 0.45, 0.73}
\definecolor{gray}{rgb}{0.5,0.5,0.5} 
\definecolor{green}{rgb}{0, 0.4, 0} 
\definecolor{orange}{rgb}{1, 0.5, 0} 	
\definecolor{mahogany}{rgb}{0.75, 0.25, 0.0}
\definecolor{purple}{rgb}{0.6, 0, 0.6}
\definecolor{darkgreen}{rgb}{0, 0.4, 0.4} 
\definecolor{teal}{rgb}{0.0, 0.5, 0.5}
\definecolor{aaaa}{rgb}{0.55, 0.1, 0.7}
\definecolor{red}{rgb}{1.0, 0, 0}

\newboolean{revising}
\setboolean{revising}{true}
\ifthenelse{\boolean{revising}}
{
	\newcommand{\fuen}[1]{\textcolor{blue}{[FuEn]: #1}}

} {
	\newcommand{\fuen}[1]{#1}

}
\newcommand{\DDR}{Differentiable Depth Rendering}
\newcommand{\HD}{horizon-depth}

\def\cvprPaperID{800} 
\def\confYear{CVPR 2021}

\begin{document}

\title{LED{\boldmath$^2$}-Net: Monocular 360{\boldmath$^\circ$} Layout Estimation via \DDR}

\author{
    Fu-En Wang$^{* 1}$ \\
    {\tt\small fulton84717@gapp.nthu.edu.tw}
    \and
    Yu-Hsuan Yeh$^{* 2}$\\
    {\tt\small yuhsuan.eic08g@nctu.edu.tw}
    \and
    Min Sun$^{1,4}$\\
    {\tt\small sunmin@ee.nthu.edu.tw}
    \and
    Wei-Chen Chiu$^{2}$\\
    {\tt\small walon@cs.nctu.edu.tw}
    \and
    Yi-Hsuan Tsai$^{3}$\\
    {\tt\small ytsai@nec-labs.com} 
}

\maketitle
\setcounter{page}{1}
\begin{abstract}
Although significant progress has been made in room layout estimation, most methods aim to reduce the loss in the 2D pixel coordinate rather than exploiting the room structure in the 3D space. Towards reconstructing the room layout in 3D, we formulate the task of 360$^\circ$ layout estimation as a problem of predicting depth on the horizon line of a panorama. Specifically, we propose the {\DDR} procedure to make the conversion from layout to depth prediction differentiable, thus making our proposed model end-to-end trainable while leveraging the 3D geometric information, without the need of providing the ground truth depth. Our method achieves state-of-the-art performance on numerous 360$^\circ$ layout benchmark datasets. Moreover, our formulation enables a pre-training step on the depth dataset, which further improves the generalizability of our layout estimation model.
\end{abstract}

\myfootnote{1}{National Tsing Hua University}
\myfootnote{2}{National Chiao Tung University }
\myfootnote{3}{NEC Labs America}
\myfootnote{4}{\scriptsize MOST Joint Research Center for AI Technology and All Vista Healthcare}
\myfootnote{*}{The authors contribute equally to this paper.}

\section{Introduction}
\label{sec:intro}
Inferring the geometric structure such as depth, layout, etc. from a single image has been studied for years. With the advance of deep learning, convolutional neural networks are widely used in these tasks. In addition, with the increasing popularity of consumer-level 360$^\circ$ cameras, approaches dealing with 360$^\circ$ panoramas start to play a crucial role in virtual and augmented reality (VR/AR) and robotic vision. In order to support the indoor use case of these applications, the task of room layout estimation from a single 360$^\circ$ panorama becomes important.

\begin{figure}[thbp]
    \centering
    \includegraphics[width=\columnwidth]{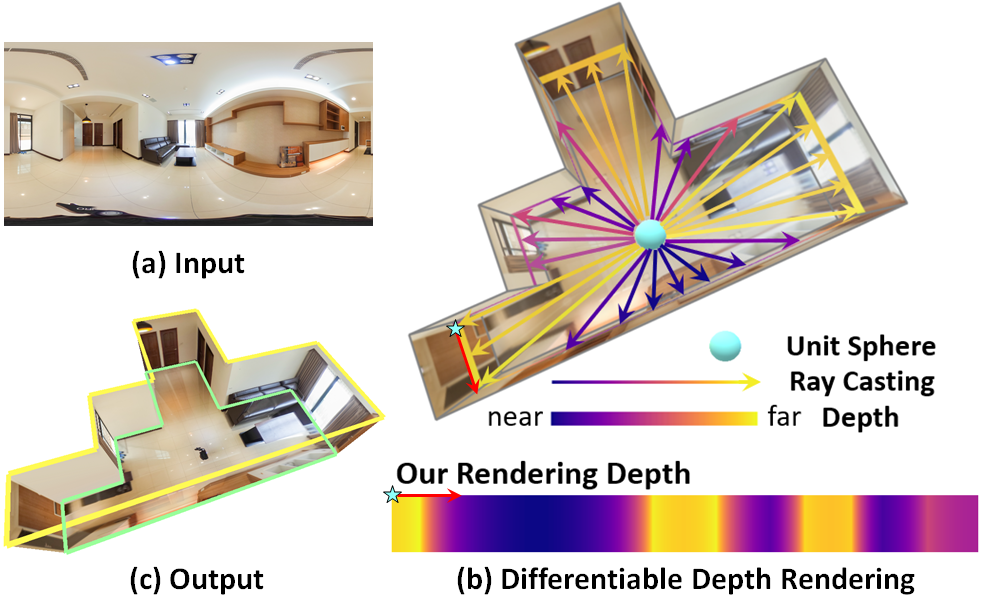}
    \caption{Our LED$^2$-Net takes the (a) single panorama as input and infers the (c) 3D room layout. We propose the (b) Differentiable Depth Rendering technique to incorporate the geometry-aware information into our model.}
    \label{fig:teaser}
    \vspace{-3mm}
\end{figure}

\begin{figure}[t]
    \centering
    \includegraphics[width=\columnwidth]{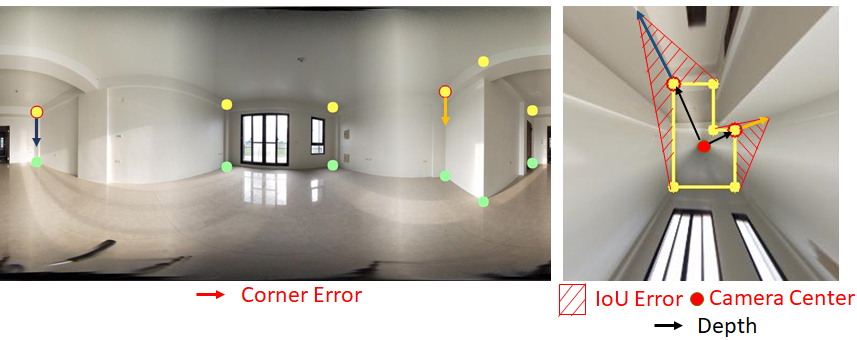}
    \caption{
    For the panorama shown on the left, we visualize several corners/boundary points where the layout estimation methods generally aim to find, in which their objective is mostly based on the errors in the 2D pixel coordinate on the equirectangular image (e.g., two arrows indicate the same error).
    However, as illustrated in the ceiling perspective view on the right, these two corner errors actually associate with different depth values (denoted as black arrows) from the camera, and such depth difference cannot be reflected in the intersection-over-union (IoU) error metric.
    }
    \label{fig:motivation}
    \vspace{-3mm}
\end{figure}

Generally, the room layout can be constructed by connecting the adjacent room corners or directly finding the boundary between walls, floor, and ceiling. Hence, most methods directly estimate the layout boundary and corners from the input panorama, e.g., HorizonNet~\cite{horizonnet}. 
Despite significant progress being achieved, the 3D reconstruction of the room layout is often not as good as expected from observing the results overlaid on the 2D panorama.
The main issue is that these methods are trained with the loss in the pixel coordinates of the 2D panorama rather than in the coordinate of the 3D reconstruction. In particular, 2D pixel loss disregards the fact that pixels with different depths from the camera should contribute differently to the loss in the 3D coordinate (see Figure~\ref{fig:motivation}). Additional losses such as binary segmentation loss in the ceiling and floor perspective views have been introduced~\cite{dula-net}. However, segmentation loss tends to focus on the correctness of the majority of the segment rather than the boundary of the segment.
%
On the other hand, although several progresses have been made for monocular 360$^\circ$ depth estimation given a single 2D panorama \cite{360depth_layout_cvpr, 360depth_layout_eccv, bifuse, omnidepth} where the loss is defined to reduce errors in 3D, none of the existing works aims at applying depth-based constraints to layout estimation framework. Hence, we are inspired to leverage the depth prediction loss to improve room layout estimation, which provides us the geometric information in the 3D space.

\begin{figure}[t]
    \centering
    \includegraphics[width=\columnwidth]{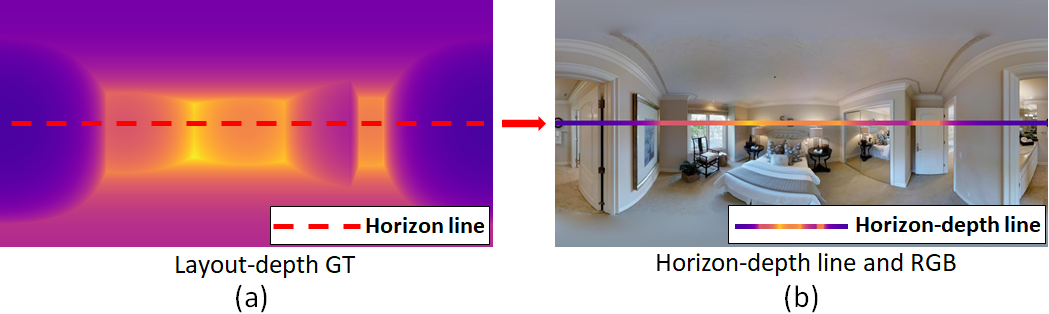}
    \caption{(a) The layout-depth generated from layout annotation. The horizontal red line indicates the {\HD}, in which we use it as the supervisory signal for the network. (b) The {\HD} aligned with the RGB panorama.}
    \label{fig:layout-depth}
    \vspace{-2mm}
\end{figure}

To this end, we re-formulate the 360$^\circ$ layout estimation into a unique 360$^\circ$ depth estimation problem.
First, instead of trying to estimate the full depth map of the panorama, we only estimate the depth values on the horizon line of a panorama, which we call ``\HD'' (see Figure~\ref{fig:layout-depth}), which is already sufficient to recover the layout.
To this end, we propose a differentiable \textit{L2D} (Layout-to-Depth) procedure to transform the layout into a ``layout-depth''.
As a result, we can adopt the widely-used objective functions in depth estimation to train our model on layout estimation datasets, which also enables the possibility to pre-train our model on depth estimation datasets and further improve the model generalizability.

Our proposed layout-to-depth procedure is based on ray-casting (i.e., casting the rays from a unit sphere as illustrated in Figure \ref{fig:teaser}(b)). The depth is recovered by computing the distances of each ray. Ideally, we can predict the depth for every pixel on the horizon line, but it would reduce the model efficiency. Also, for layout estimation, we simply need to know at least the depth values of corner points in the room. To consider the balance between efficiency and accuracy, we propose a ``Grid Re-sample'' strategy which is able to approximate the {\HD} map by a flexible number of casting rays (see Figure~\ref{fig:teaser}(b)).
We name our method \textbf{LED{\boldmath$^2$}-Net}, which can be efficiently trained in an end-to-end fashion.

To demonstrate the effectiveness of our proposed model based on the novel technique of {\DDR} for 360$^\circ$ layout estimation, we conduct extensive experiments on four benchmark datasets, including Matterport3D~\cite{jw_layout}, Realtor360~\cite{dula-net}, PanoContext~\cite{panocontext}, and Stanford2D3D~\cite{stfd}.
We show that our method performs favorably against state-of-the-art approaches in both the within-dataset and cross-dataset settings.
More interestingly, we leverage the property of our depth estimation objective to enable depth pre-training using a synthetic dataset, Structured3D~\cite{structure3D}, which further improves the generalization ability of our model. Our supplementary material, source code, and pre-trained models are available to the public on our project website\footnote{\url{https://fuenwang.ml/project/led2net}}.
We summarize our contributions as follows:
\begin{enumerate}
    \item We reformulate the task of 360$^\circ$ layout estimation to a unique 360$^\circ$ depth estimation problem that optimizes a loss in 3D while maintaining the simplicity of layout estimation.
    \item We propose a differentiable layout-to-depth procedure to convert the layout into horizon-depth through ray-casting of a few points, which enables the end-to-end training on layout estimation datasets.
    \item We show that our framework can be seamlessly pre-trained by 360$^\circ$ depth datasets, which further improves the generalizability on cross-dataset evaluations.
\end{enumerate}

\section{Related Work}
With the popularity of virtual/augmented reality, inferring the geometric context from 360$^\circ$ images becomes an important topic in recent years. In this section, we discuss the literature relevant to 360$^\circ$ depth and layout estimation.

\vspace{-3mm}
\paragraph{360{\boldmath$^\circ$} Depth and Layout Estimation.} 
One of the pioneering works for 360$^\circ$ perception is proposed by Cheng~\etal~\cite{cubepadding}. They use cubemap projection and cube padding to avoid the distortion of equirectangular images while keeping the connection between each adjacent face of the cubemap. Wang~\etal~\cite{ouraccv} then adopt the cubemap representation and unsupervisedly learn monocular 360$^\circ$ depth estimation. To capture distortion-aware context, several approaches of spherical CNNs are proposed~\cite{yu-sphericalcnn, KTN, sphericalcnns, Yang_2020_CVPR, Coors_2018_ECCV, Esteves_2018_ECCV}. 
With a supervised scheme, Zioulis~\etal~\cite{omnidepth} incorporate \cite{yu-sphericalcnn} and propose two network variants to estimate monocular 360$^\circ$ depth. Following \cite{ouraccv, omnidepth}, Wang~\etal~\cite{bifuse} propose a framework consisting of equirectangular and cubemap projections to estimate the 360$^\circ$ depth map.

While the depth map contains details of a scene, the layout provides a rough room structure. Since most rooms have walls that are perpendicular to each other, many approaches follow the assumption of Manhattan World~\cite{manhattan}. By using the Line Segment Detection~\cite{LSD} and extracting the vanishing points, Zhang~\etal~\cite{panocontext} generate the layout hypothesis and infers the layout from a single 360$^\circ$ panorama. Recently, CNN-based approaches have come in handy for layout estimation. Zou~\etal~\cite{layoutnet} first incorporate a U-Net-like~\cite{UNet} architecture to predict the corners/boundary of the room, and then apply the optimization based on the Manhattan assumption. Sun~\etal~\cite{horizonnet} propose HorizonNet which simplifies the layout into a horizontal representation and improves it via a recurrent neural network.
On the other hand, as the bird's eye view of the layout can be considered as the floor plan, several approaches consider this property and instead infer the floor plan of a room. Yang~\etal~\cite{dula-net} propose DuLa-Net to predict a binary segmentation map as the floor plan. Pintore~\etal~\cite{atlantanet} predict the floor plan by combining the benefits of both DuLa-Net and HorizonNet. 

Different from the above-mentioned approaches, we find that the geometric cues across the layout and depth are tightly relevant to each other, and combining the two information becomes an important topic recently. Jin~\etal~\cite{360depth_layout_cvpr} propose to use the layout information (i.e. boundary, corners, and depth) to improve 360$^\circ$ depth estimation. Zeng~\etal~\cite{360depth_layout_eccv} propose a two-stage framework to estimate both depth and layout-depth. However, since the annotation of layout involves only monocular images, the room scale is unknown. Thus, direct regression for up-to-scale layout-depth suffers from unknown scale issues. This motivates us to design a representation which is differentiable, geometric-aware, scale-invariant, and efficient to optimize, in a way that approximates the dense depth map.

\section{Approach}
\begin{figure*}[thbp]
    \centering
    \includegraphics[width=\textwidth]{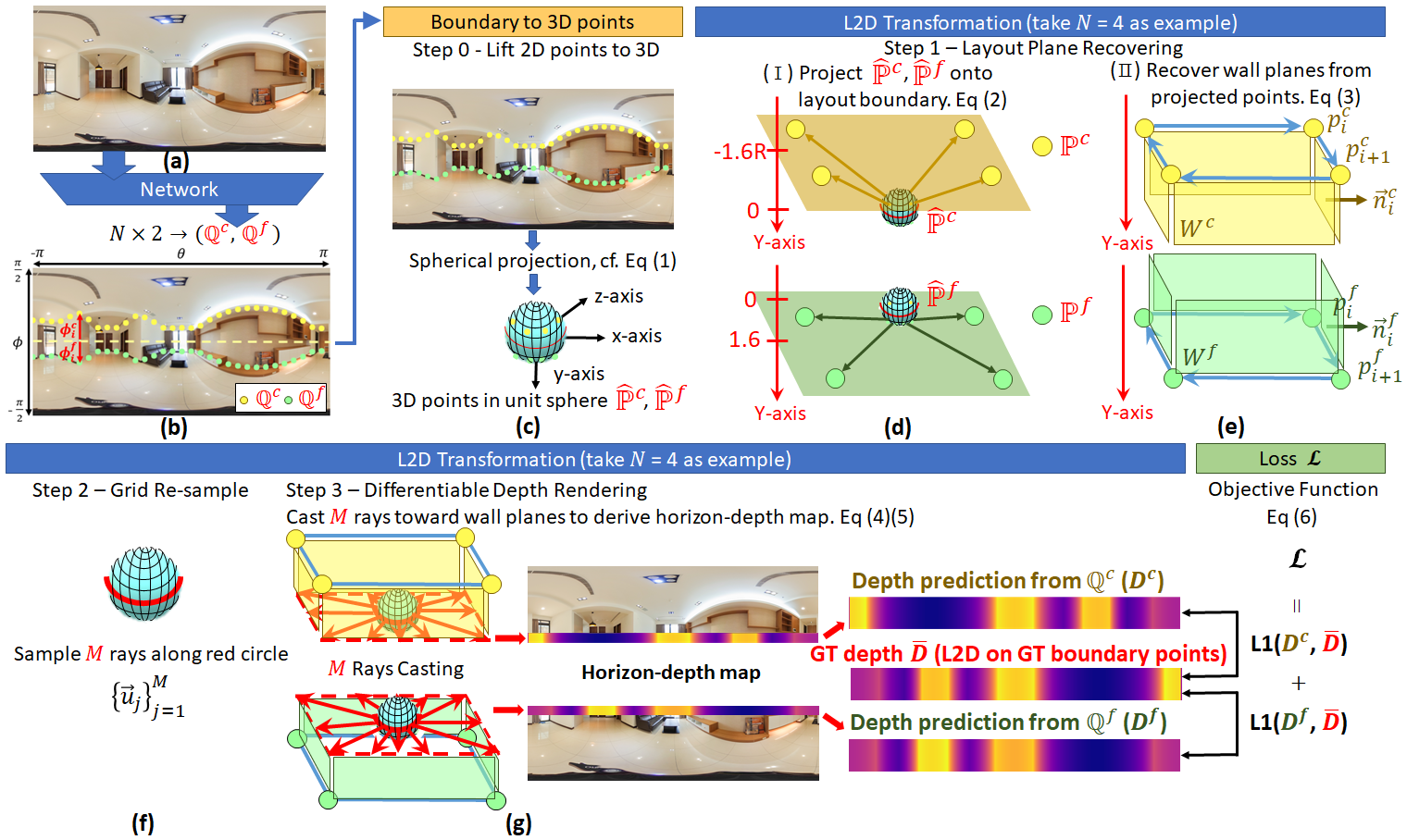}
    \caption{
    The overall framework of our proposed LED$^2$-Net. Our layout estimation network takes \textbf{(a)} an RGB panorama as input, and outputs \textbf{(b)} the layout representation which is composed of two sets of layout boundary points ($\mathbb{Q}^c, \mathbb{Q}^f$) on the ceiling and the floor respectively. Our novel L2D (layout-to-depth) transformation is applied to transform the layout representation, which is first \textbf{(c)} lifted to 3D space (cf. Section~\ref{sec:preliminary}), into the {\HD} maps (i.e. $D^c$ and $D^f$). The L2D transformation is composed of three main steps: \textbf{(d)(e)}``Layout Plane Recovering'' (cf. Section~\ref{Plane-Re}) to recover the plane equation (i.e. $W^c$ and $W^f$), \textbf{(f)} ``Grid Re-sample'' (cf. Section~\ref{Plane-Re}), and \textbf{(g)}``\DDR'' (cf. Section~\ref{DDR}). The training objective of our LED$^2$-Net is thus defined by the errors between our estimated {\HD} maps and the corresponding ground truth $\bar{D}$ (cf. Section~\ref{sec:loss}).
    }
    \label{fig:arch}
\end{figure*}

As motivated previously, in this paper we aim to use the layout-depth as the training objective for 360$^\circ$ layout estimation, which is realized by our ``L2D'' (Layout-to-Depth) transformation. 
Basically, our layout estimation network takes a panorama as input and learns to predict the spherical coordinates of boundary points on the equirectangular image. Note that boundary points outlining both the floor and ceiling are estimated, as shown in Figure~\ref{fig:arch}(b).
Afterward, the L2D transformation is applied to the predicted boundary points to establish the {\HD} map, as shown in Figure~\ref{fig:arch}(c-g). The errors on such a {\HD} map with respect to the ground truth become the objective for training our layout estimation network.
In the following, we first introduce the layout representation and spherical projection in Section~\ref{sec:preliminary}, followed by elaborating the L2D transformation in Section~\ref{sec:l2d}. Finally, we provide details of our loss function and network architecture in Section~\ref{sec:loss} and Section~\ref{sec:network}, respectively.


\subsection{Preliminary} 
\label{sec:preliminary}

Given an equirectangular image taken in a room, we represent its layout with a sequence of boundary positions in the spherical coordinate system, as shown in Figure~\ref{fig:arch}(b). Basically, a pixel $q$ on the equirectangular image positioned by longitude and latitude, i.e., $(\theta, \phi)$, can be easily converted to a 3D point $p \in \mathbb{R}^3$ on a unit sphere by the function $S(\theta, ~\phi)$, as shown in Figure~\ref{fig:arch}(c):
\begin{align}   
    \begin{split}
    S(&\theta, ~\phi) = (p_x, ~p_y, ~p_z)~, \\
    p_x &= \cos(\phi) \cdot \sin(\theta)~, \\
    p_y &= \sin(\phi)~, \\
    p_z &= \cos(\phi) \cdot \cos(\theta)~.
    \end{split}
    \label{eq:spherical-projection-forward}
\end{align}
As $\theta$ and $\phi$ indicate a location on the sphere, the range of $\theta$ is from $-\pi$ to $+\pi$, while the range of $\phi$ is from $-0.5\pi$ to $+0.5\pi$. In the following, we use such function $S$ for converting the layout boundary to 3D points on a unit sphere.

\subsection{L2D Transformation}
\label{sec:l2d}
Before we delve into the details of our proposed L2D transformation, we first introduce two important assumptions used in most prior works of 360$^\circ$ layout estimation, where these two assumptions help to tackle the absolute scale issue that typically cannot be derived from a monocular image:
\begin{enumerate}
    \item The height of the camera for taking panoramas (i.e., the perpendicular distance from the floor) is normalized to a fixed number~\cite{dula-net, horizonnet}. Following \cite{horizonnet}, it is set to 1.6 for all the experiments in this paper.
    \item The ceilings, floors, and walls are flat planes, where the walls are perpendicular to each other (i.e., Manhattan World assumption~\cite{manhattan}).
\end{enumerate}
Our proposed method also follows these two assumptions for transforming the layout into depth prediction. In addition, the two assumptions are applied to generate the ground truth {\HD}.
In the following, we introduce three main steps of our L2D transformation.

\vspace{-3mm}
\paragraph{Layout Plane Recovering.}
\label{Plane-Re}
Our layout representation predicted from the input panorama is composed of two sets of spherical coordinates, denoted respectively as $\mathbb{Q}^f$ and $\mathbb{Q}^c$, where $\mathbb{Q}^f = \{q^f_i\}_{i=1}^N,~q^f_i = (\theta^f_i, \phi^f_i)$ and $\mathbb{Q}^c = \{q^c_i\}_{i=1}^N,~q^c_i = (\theta^c_i, \phi^c_i)$. The set $\mathbb{Q}^f$ (respectively $\mathbb{Q}^c$) represents the boundary points 
sampled from the boundaries between the walls and the floor (respectively the ceiling), where $N$ is the number of boundary points and the smallest $N$ is equal to the number of walls.
Particularly, $q^f_i$ and $q^c_j$ with $i = j$ share the same value of longitude $\theta$. Note that, the points in $\mathbb{Q}^f$ or $\mathbb{Q}^c$ are already arranged by the ascending order according to their values of longitude $\theta$.
We can convert these two point sets from spherical coordinate system onto a unit sphere by using function $S$ via \eqref{eq:spherical-projection-forward}, and obtain $\hat{\mathbb{P}}^f = \{\hat{p}^f_i=S(q^f_i)\}_{i=1}^N$ and $\hat{\mathbb{P}}^c = \{\hat{p}^c_i=S(q^c_i)\}_{i=1}^N$, where each $\hat{p}$ is a 3-dimensional vector in the Cartesian coordinate system with $\norm{\hat{p}}=1$. This step is illustrated in Figure~\ref{fig:arch}(c).

Next, based on the two aforementioned assumptions of Manhattan layout, i.e., the camera height is fixed to 1.6 and the floor (respectively the ceiling) is a flat plane, we project all the points in $\hat{\mathbb{P}}^f$ (respectively $\hat{\mathbb{P}}^c$) from the unit sphere onto the boundary between the walls and the floor (respectively the ceiling):
\begin{equation}
    \begin{aligned}
    p_i^f & = \hat{p}_{i}^f * \frac{1.6}{\hat{p}_{i}^f(y)}, \\
    p_i^c & = \hat{p}_{i}^c * \frac{-1.6R}{\hat{p}_{i}^c(y)},
    \end{aligned}
    \label{eq:ceiling-floor-normalize}
\end{equation}
where $\hat{p}_{i}^f(y)$ denotes the coordinate of $\hat{p}_{i}^f$ in the $y-$axis, and similarly for $\hat{p}_{i}^c(y)$. $R$ denotes the ratio between the height of the camera and the distance from the camera center to the ceiling.
Figure~\ref{fig:arch}(d) presents the step of this projection. Note that the $y-$axis in this Cartesian coordinate system is perpendicular to the ground plane and point to the floor.

After obtaining $\mathbb{P}^f = \{p^f_i\}^N_{i=1}$ and $\mathbb{P}^c = \{p^c_i\}^N_{i=1}$, based on every pair of adjacent points in $\mathbb{P}^f$, we derive $N$ walls where their plane equations $\{W_i^f\}^N_{i=1}$ are obtained by:
\begin{equation}
    \begin{aligned}
    \vec{n}^f_i &= \hat{y} \times (p^f_{i+1} - p^f_i)~,\\
    t^f_i &= -\vec{n}^f_i \cdot p^f_i~,\\
    W^f_i &= (\vec{n}^f_i, t^f_i)~.
    \end{aligned}
    \label{eq:plane_equations}
\end{equation}
where $\times$ denotes the outer product operation, $\hat{y}$ denotes the unit vector along $y-$axis, $\vec{n}$ is the 3-dimensional normal vector of the wall, and $t^f_i$ is the offset in the plane equation. This plane recovering step is shown in Figure~\ref{fig:arch}(e).
Another set of walls based on $\mathbb{P}^c$ with plane equations $\{W_i^c\}^N_{i=1}$  can be also derived in the same way.

\vspace{-3mm}
\paragraph{Grid Re-sample.}
\label{Grid-Re}
After having the wall equations in the ``layout plane recovering'' step, we aim at casting the rays from the unit sphere (i.e., camera center) towards these walls in order to obtain the information related to depth. As motivated in the introduction, we take the computational efficiency into consideration, in which the number of casting rays is less than the size of the complete horizon-depth map (i.e., covering the width of the input panorama). In other words, the {\HD} map generated by our ray-casting process is an approximation of the complete one.

Here, we approximate the {\HD} map with a size $M$ by sampling $M$ rays casting from the unit sphere (see Figure~\ref{fig:arch}(f)), in which these rays are denoted as $M$ unit vectors $\{\vec{u}_j\}^M_{j=1}~, \norm{\vec{u}_j} = 1$.
Specifically, these rays are obtained by $\vec{u}_j = S(\theta_j, \phi_j),~j=\{1,..., M\}$, where $\phi_j=0~\forall j$ and $\{\theta_j\}_{j=1}^M$ are equiangularly sampled from $[-\pi, \pi]$. That is, these rays form a 360$^\circ$ horizontal radiation pattern. Note that, the zero latitude is aligned with the height of the camera, which is a general setting in the prior work~\cite{horizonnet, dula-net}.

\vspace{-3mm}
\paragraph{\DDR.}\label{DDR}
Now, as we already have the wall planes and the casting rays, we can compute the intersections of them, and then the {\HD} map can be easily obtained from the distances between these intersections and the camera center. To be detailed, given wall planes $\{W^f_i\}^N_{i=1}$ and casting rays $\{\vec{u}_j\}^M_{j=1}$, for each $\vec{u}_j$, we obtain $N$ candidate depth values $\{d_{j, i}^f\}^N_{i=1}$ as:
\begin{equation}
    d_{j, i}^f =-\frac{t^f_i}{\vec{u}_j \cdot \vec{n}^f_i}.
\end{equation}
In particular, we use two conditions to filter out inappropriate candidates: (1) $d_{j, i}^f$ must be $\geq 0$;
and (2) since the wall $W^f_i$ is derived from $p^f_{i+1}$ and $p^f_i$ via \eqref{eq:plane_equations}, which are connected to the longitude $\theta^f_{i+1}$ and $\theta^f_i$, the longitude of the intersection of $\vec{u}_j$ and $W^f_i$ must be within the range $[\theta^f_i, \theta^f_{i+1}]$.
After filtering out the candidates, we obtain the corresponding depth value $d_{j}^f$ of the $j-$th pixel on the resultant {\HD} map of size $M$ via:
\begin{equation}
d_{j}^f = \min_{i} d_{j, i}^f,
\end{equation}
where we use the $\min$ function to find $d_{j}^f$, since other larger values indicate the occluded area. The same computation procedure can be applied to another set of wall planes $\{W^c_i\}_{i=1}^N$. We denote the final {\HD} maps by concatenating $d_{j}^f$ (respectively $d_{j}^c$) as $D^f$ (respectively $D^c$) in Figure~\ref{fig:arch}(g).

\subsection{Objective Function}
\begin{table*}[htbp]
  \centering
  \caption{The quantitative experimental results on Realtor360~\cite{dula-net} dataset.}
  \resizebox{\textwidth}{!}{
    \begin{tabular}{|c|c|c|c|c|c|c|c|c|c|c|}
    \hline
    \multirow{2}[4]{*}{Method} & \multicolumn{2}{c|}{Overall} & \multicolumn{2}{c|}{4 corners} & \multicolumn{2}{c|}{6 corenrs} & \multicolumn{2}{c|}{8 corners} & \multicolumn{2}{c|}{10+ corners} \bigstrut\\
    \cline{2-11}          & 2D IoU (\%) & 3D IoU (\%) & 2D IoU (\%) & 3D IoU (\%) & 2D IoU (\%) & 3D IoU (\%) & 2D IoU (\%) & 3D IoU (\%) & 2D IoU (\%) & 3D IoU (\%) \bigstrut\\
    \hline
    LayoutNet & 65.84 & 62.77 & 80.41 & 76.60 & 60.50 & 57.87 & 41.16 & 41.16 & 22.35 & 22.35 \bigstrut[t]\\
    DuLa-Net & 80.53 & 77.20 & 82.63 & 78.91 & 80.72 & 77.79 & 78.12 & 74.86 & 63.10 & 59.72 \\
    HorizonNet & 86.69 & 83.66 & 87.83 & 84.73 & 87.63 & 84.78 & 81.27 & 78.44 & 78.49 & 73.64 \\
    AtlantaNet & 80.36 & 74.59 & 83.42 & 77.05 & 80.67 & 75.01 & 73.72 & 69.31 & 59.43 & 55.51 \\
    \textbf{Ours} & \textbf{88.19} & \textbf{85.21} & \textbf{89.25} & \textbf{86.33} & \textbf{88.80} & \textbf{85.97} & \textbf{83.70} & \textbf{80.81} & \textbf{81.67} & \textbf{76.20} \bigstrut[b]\\
    \hline
    \end{tabular}%
  }
  \label{tab:realtor360}%
\end{table*}%
\begin{table*}[htbp]
  \centering
  \caption{The quantitative experimental results on Mattertport3D~\cite{jw_layout} dataset.}
  \resizebox{\textwidth}{!}{
    \begin{tabular}{|c|c|c|c|c|c|c|c|c|c|c|}
    \hline
    \multirow{2}[4]{*}{Method} & \multicolumn{2}{c|}{Overall} & \multicolumn{2}{c|}{4 corners} & \multicolumn{2}{c|}{6 corenrs} & \multicolumn{2}{c|}{8 corners} & \multicolumn{2}{c|}{10+ corners} \bigstrut\\
\cline{2-11}          & \multicolumn{1}{c|}{2D IoU (\%)} & \multicolumn{1}{c|}{3D IoU (\%)} & \multicolumn{1}{c|}{2D IoU (\%)} & \multicolumn{1}{c|}{3D IoU (\%)} & \multicolumn{1}{c|}{2D IoU (\%)} & \multicolumn{1}{c|}{3D IoU (\%)} & \multicolumn{1}{c|}{2D IoU (\%)} & \multicolumn{1}{c|}{3D IoU (\%)} & \multicolumn{1}{c|}{2D IoU (\%)} & \multicolumn{1}{c|}{3D IoU (\%)} \bigstrut\\
    \hline
    LayoutNet & 78.73 & 75.82 & 84.61 & 81.35 & 75.02 & 72.33 & 69.79 & 67.45 & 65.14 & 63.00 \bigstrut[t]\\
    DuLa-Net & 78.82 & 75.05 & 81.12 & 77.02 & 82.69 & 78.79 & 74.00 & 71.03 & 66.12 & 63.27 \\
    HorizonNet & 81.24 & 78.73 & 83.54 & 80.81 & 82.91 & 80.61 & 76.26 & 74.10 & 72.47 & 70.30 \\
    AtlantaNet & 82.09 & 80.02 & 84.42 & 82.09 & 83.85 & 82.08 & 76.97 & 75.19 & \textbf{73.19} & \textbf{71.62} \\
    \textbf{Ours} & \textbf{83.91} & \textbf{81.52} & \textbf{86.91} & \textbf{84.22} & \textbf{85.53} & \textbf{83.22} & \textbf{78.72} & \textbf{76.89} & 71.79 & 70.09 \bigstrut[b]\\
    \hline
    \end{tabular}%
  }
  \label{tab:mp3djw}%
  \vspace{-3mm}
\end{table*}%

\begin{table}[thbp]
  \centering
  \caption{The qualitative experimental results on both PanoContext~\cite{panocontext} and Stanford2D3D~\cite{stfd} datasets.}
    \resizebox{\columnwidth}{!}{\begin{tabular}{|c|c|c|c|c|c|}
    \hline
    \multirow{2}[4]{*}{Method} & \multicolumn{5}{c|}{3D IoU (\%)} \bigstrut\\
\cline{2-6}          & LayoutNet & DuLa-Net & HorizonNet & AtlantaNet & \textbf{Ours} \bigstrut\\
    \hline
    PanoContext & 74.48 & 77.42 & 82.17 & 78.76 & \textbf{82.75} \bigstrut[t]\\
    Stanford2D3D & 76.33 & 79.36 & 79.79 & 82.43 & \textbf{83.77} \bigstrut[b]\\
    \hline
    \end{tabular}}
  \label{tab:cuboid}%
  \vspace{-2mm}
\end{table}%

\label{sec:loss}
As the entire procedure of our proposed L2D (layout-to-depth) transformation is differentiable, the training objective of our model can be directly defined upon the errors of our two predicted {\HD} maps (i.e., $D^f$ and $D^c$) with respect to the ground truth {\HD} map $\bar{D}$. Moreover, the model is end-to-end trainable, where we adopt the L1 loss to measure the errors between depth maps:
\begin{equation}
    \mathcal{L} = \norm{D^f - \bar{D}}_1 + \norm{D^c - \bar{D}}_1.
    \label{eq:loss}
\end{equation}
It is worth noting that, the ground truth depth map $\bar{D}$ can be obtained via applying the same L2D transformation on the ground truth layout, which consists of the ground truth layout boundary points. Also, as the depth maps obtained from the ground truth boundary points either on the floor or the ceiling should be identical to each other, here we only use a single ground truth {\HD} map $\bar{D}$.
However, using the depth objective may encounter the scaling issue caused by the unknown scale in a room. In our method, since the point sets $\mathbb{Q}^f$ and $\mathbb{Q}^c$ only represent the angles, which are scale-invariant, this effect does not exist anymore.

Another benefit of our learning objective based on a depth loss is that we can not only use the {\HD} map derived from the layout, but also the ones acquired from the laser scanner or virtual environments. Later in our experimental section, we demonstrate that using a virtual 360$^\circ$ dataset with depth ground truths to pre-train our layout estimation network can further improve its generalization ability on the cross-dataset evaluation setting.

\subsection{Network Architecture}
\label{sec:network}
We follow the architecture of the HorizonNet~\cite{horizonnet} to construct our layout estimation network.
First, a ResNet-50~\cite{ResNet} based encoder is adopted to extract feature maps of the input equirectangular image, where the feature maps at different scales are further fused together via several convolution layers followed by concatenation~\cite{FPN}.
Then, a bidirectional Long Short-Term Memory (bi-LSTM) module~\cite{biRNN, lstm} is applied to smooth the fused feature map along the width, followed by one fully-connected layer and a sigmoid function to obtain our final output.
In our formulation, the layout representation estimated by our network is composed of two point sets in the spherical coordinate, i.e., $\mathbb{Q}^f$ and $\mathbb{Q}^c$. In our implementation, since we distribute the $N$ boundary points along the axis of longitude sampled from $[-\pi, \pi]$, our layout estimation network only needs to predict the latitude values of the boundary points.
Thus, the second dimension of the output size $N\times 2$ indicates the two point sets related to the ceiling and the floor ($N$ is set to 256 in our experiments).
To further constrain the layout prediction by the Manhattan assumption and infer the layout height to create a clean room layout, we adopt the post processing procedure of HorizonNet~\cite{horizonnet}.
Please refer to the supplementary material for more detailed descriptions of our network architecture and the post processing step.

\section{Experiments}

We conduct extensive experiments on four 360$^\circ$ layout datasets, which are Realtor360~\cite{dula-net} and Matterport3D~\cite{jw_layout} with more complicated scenes, and two cuboid datasets, PanoContext~\cite{panocontext} and Stanford2D3D~\cite{stfd}.
We compare our proposed method with several state-of-the-art baselines of monocular 360$^\circ$ layout estimation, including LayoutNet~\cite{layoutnet}, DuLa-Net~\cite{dula-net}, HorizonNet~\cite{horizonnet}, and AtlantaNet~\cite{atlantanet}. Moreover, since our method allows us to use the datasets that contain the depth ground truth as a pre-training step, we conduct additional experiments by leveraging a synthetic dataset (i.e., Structured3D~\cite{structure3D}), in which the depth annotation is free to collect.
Note that such pre-training is an additional benefit but not the requirement of our framework, in which none of the existing layout estimation methods is equipped with this ability.
We follow the same protocols of \cite{dula-net} to calculate the 2D and 3D intersection-over-union (IoU). We also investigate the model sensitivity on the ray-casting number (i.e., $M$). More results are provided in the supplementary material.

\begin{figure*}[thbp]
    \centering
    \includegraphics[width=\textwidth]{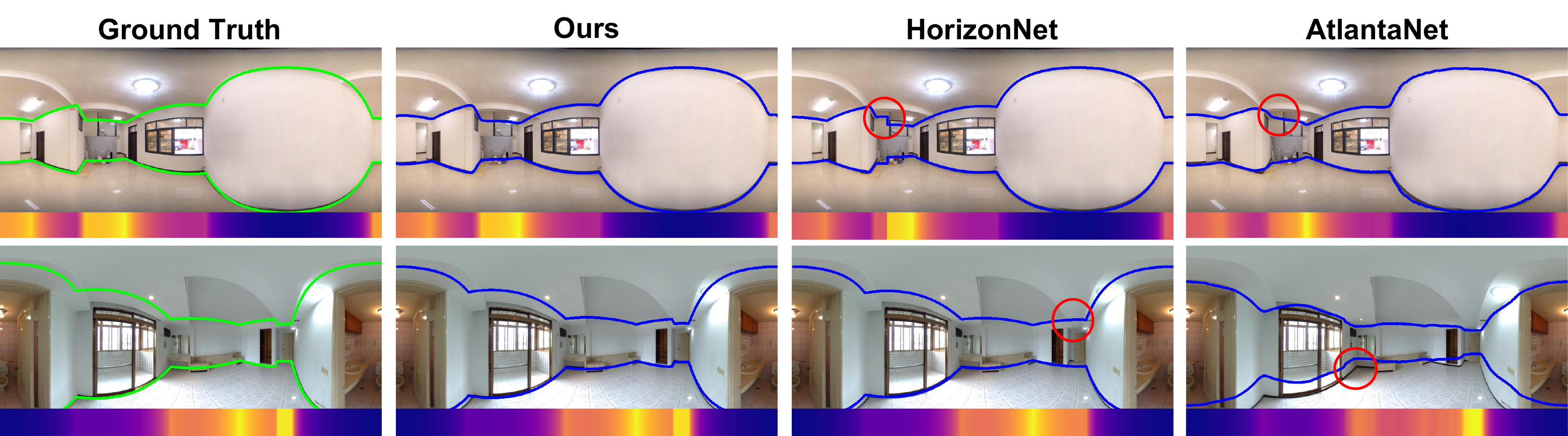}
    \caption{The qualitative results of layout boundary and {\HD} on the Realtor360~\cite{dula-net} dataset. The red circles highlight the errors produced by the baselines.}
    \label{fig:realtor360}
    \vspace{-3mm}
\end{figure*}
\begin{figure*}[thbp]
    \centering
    \includegraphics[width=\textwidth]{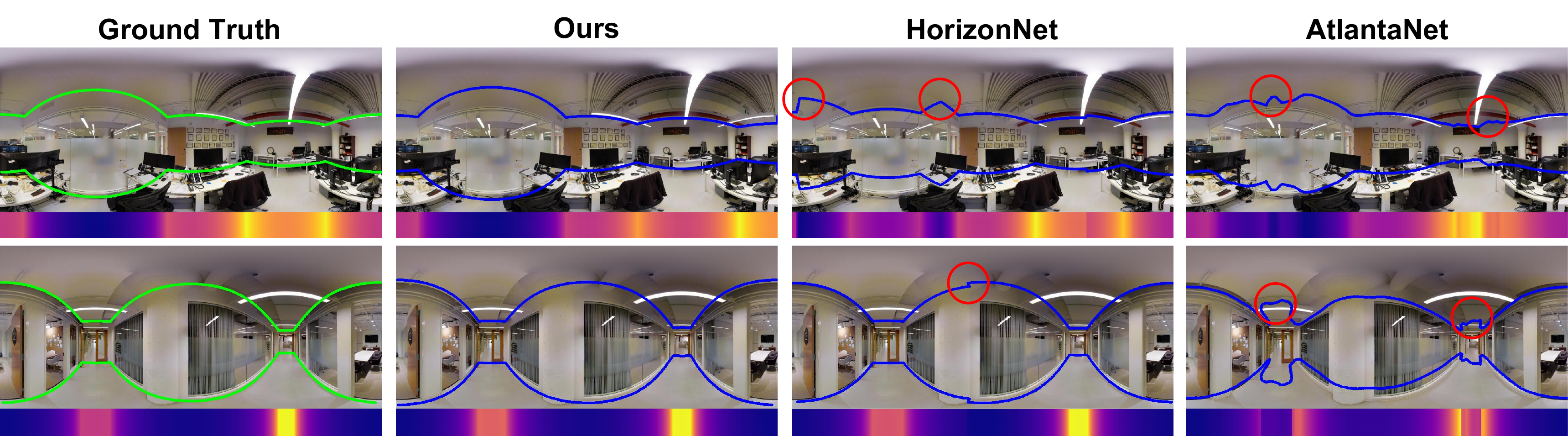}
    \caption{The qualitative results of layout boundary and {\HD} on the Matterport3D~\cite{jw_layout} dataset. The red circles highlight the errors produced by the baselines.}
    \label{fig:mp3djw}
    \vspace{-3mm}
\end{figure*}
\begin{table*}[t]
  \centering
  \caption{The quantitative results of the cross-dataset evaluation scheme (cf. Section~\ref{sec:Generalizability}).}
    \resizebox{\textwidth}{!}{\begin{tabular}{|c|c||c|c|c||c|c|c|}
    \hline
    \multirow{2}[4]{*}{Method} & \multirow{2}[4]{*}{IoU (\%)} & Train-Dataset & \multicolumn{2}{c||}{Cross-Dataset} & Train-Dataset & \multicolumn{2}{c|}{Cross-Dataset} \bigstrut\\
\cline{3-8}          &       & Matterport3D & Realtor360 & Stanford2D3D & Realtor360 & Matterport3D & Stanford2D3D \bigstrut\\
    \hline
    \hline
    \multirow{2}[2]{*}{HorizonNet} & 2D    & 81.24 & 80.01 & 84.91 & 86.69 & 78.00 & 84.84 \bigstrut[t]\\
          & 3D    & 78.73 & 76.37 & 81.74 & 83.66 & 75.24 & 80.92 \bigstrut[b]\\
    \hline
    \hline
    \multirow{2}[2]{*}{AtlantaNet} & 2D    & 73.11 & 72.26 & 81.97 & 80.36 & 73.21 & 83.48 \bigstrut[t]\\
          & 3D    & 68.09 & 66.88 & 75.22 & 74.59 & 67.76 & 77.38 \bigstrut[b]\\
    \hline
    \hline
    \multicolumn{1}{|c|}{\multirow{2}[2]{*}{Ours [w/o pretrained]}} & 2D    & \textbf{83.91} & 80.74 & 85.25 & 88.19 & 79.78 & 86.51 \bigstrut[t]\\
          & 3D    & \textbf{81.52} & 77.17 & 80.54 & 85.21 & 76.89 & 83.50 \bigstrut[b]\\
    \hline
    \hline
    \multicolumn{1}{|c|}{\multirow{2}[2]{*}{\textbf{Ours [w/ pretrained]}}} & 2D    & 83.59 & \textbf{83.73} & \textbf{88.37} & \textbf{89.00} & \textbf{79.92} & \textbf{86.81} \bigstrut[t]\\
          & 3D    & 81.24 & \textbf{80.52} & \textbf{85.20} & \textbf{86.31} & \textbf{76.99} & \textbf{83.69} \bigstrut[b]\\
    \hline
    \end{tabular}}
  \label{tab:cross-dataset}%
  \vspace{-1mm}
\end{table*}%

\begin{figure*}[thbp]
    \centering
    \includegraphics[width=0.9\textwidth]{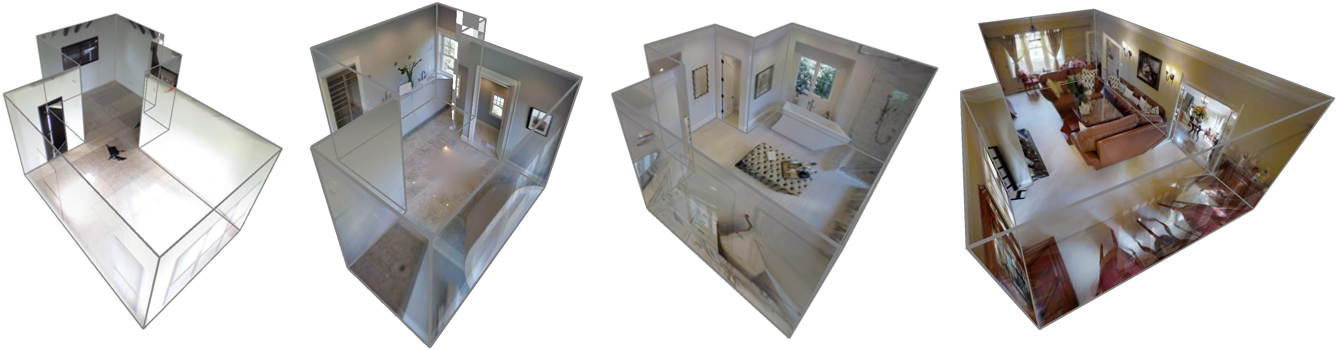}
    \caption{3D layout visualizations based on the layout estimation produced by our LED$^2$-Net, for Matterport3D and Realtor360 datasets.}
    \label{fig:more-results}
    \vspace{-2mm}
\end{figure*}
\begin{table}[htbp]
  \centering
  \caption{The ablation study for the sensitivity of our model performance with respect to the number of casting rays (i.e., $M$).}
  \vspace{2mm}
    \begin{tabular}{|c|c|c|c|c|}
    \hline
    Corner Number & 16    & 64    & 256   & 1024 \bigstrut\\
    \hline
    \hline
    2D IoU (\%) & 84.72 & 86.74 & 88.19 & 88.12 \bigstrut[t]\\
    3D IoU (\%) & 81.89 & 83.58 & 85.21 & 85.19 \bigstrut[b]\\
    \hline
    \end{tabular}%
  \label{tab:grid-resample}%
  \vspace{-3mm}
\end{table}%

\vspace{-4mm}
\paragraph{PanoContext and Stanford2D3D.} There are around 500 panoramas along with ground truth layout annotations in PanoContext dataset, which are collected from SUN360~\cite{Sun360} and labeled by Zhang~\etal~\cite{panocontext}. In order to extend the available training samples of layout estimation, Zou \etal~\cite{layoutnet} additionally collect 571 panoramas from the original Stanford2D3D dataset~\cite{stfd} and label the corresponding layout ground truths. We adopt the same train/val/test splits as used in~\cite{layoutnet} for all the experiments on these two datasets. Please note that, as PanoContext and Stanford2D3D datasets primarily consist of cuboid-shape layouts, only adopting these two datasets is not enough for well evaluating the capacity of different models for tackling layout estimation on full 360$^\circ$ panoramas. We, therefore, consider other datasets such as Realtor360 and Matterport3D, which contain more complicated cases of layouts.

\vspace{-4mm}
\paragraph{Realtor360.} 
This dataset is proposed and annotated by Yang \etal~\cite{dula-net}, where they collect 593 panoramas from the subsets of SUN360 dataset (composed of scenes of living rooms and bedrooms) as well as 1980 panoramas from a real estate database. We follow the official train/test split as~\cite{dula-net} to conduct experiments on Realtor360.


\vspace{-4mm}
\paragraph{Matterport3D.} The original Matterport3D~\cite{mp3d} dataset contains 10,800 panoramas along with the depth ground truths obtained from laser scanners. Zou~\etal~\cite{jw_layout} and Wang~\etal~\cite{layoutmp3d} remove the cases that do not satisfy the Manhattan World assumption and use the annotation tool provided by DuLa-Net~\cite{dula-net} to label the layout ground truth. Eventually, there are 2295 panoramas in total, including the complicated cases with the different number of layout corners. We adopt the official train/val/test split of \cite{jw_layout} to conduct the experiments.

\subsection{Experimental Results}

\paragraph{Datasets with Challenging Cases.} We first conduct experiments on the datasets which contain sufficiently complicated layouts (i.e., Realtor360 and Matterport3D) for making comparisons among different models in terms of their ability to deal with difficult cases. Table~\ref{tab:realtor360} and Table~\ref{tab:mp3djw} provide the quantitative results on the Realtor360 and Matterport3D datasets, respectively. On these two datasets, the proposed method performs favorably against other state-of-the-art methods. In particular, compared with HorizonNet~\cite{horizonnet} that has quite a similar architecture to our layout estimation network, our model consistently produces better performance and thus verifies the contribution of our novel L2D (layout-to-depth) transformation building upon the ``\DDR''. To further visualize the effectiveness of our approach, we provide several qualitative example results in Figure~\ref{fig:realtor360} and Figure~\ref{fig:mp3djw}, showing that our proposed method is able to infer precise layouts for complicated cases (i.e., rooms with many layout corners), while the other approaches instead produce noisy layout estimations.
In addition, we provide more 3D layout visualizations in Figure~\ref{fig:more-results}, in order to demonstrate our capability on the task of monocular 360$^\circ$ layout estimation.

\vspace{-4mm}
\paragraph{Datasets with Cuboid-shape Layouts.}
In addition to more complicated layouts, for the experiments conducted on the PanoContext and Stanford2D3D datasets, which are primarily composed of cuboid-shape layouts (i.e., the rooms have four layout corners) and considered to be simpler cases, we provide the quantitative results in Table~\ref{tab:cuboid}. We show that our proposed method consistently outperforms all the state-of-the-art methods.

\subsection{Generalizability}
\label{sec:Generalizability}

The generalizability of room layout estimation (e.g., cross-dataset setting) has not been widely studied, yet it is an important task to validate whether the models can generalize to unseen room layouts with different dataset distribution.
To investigate this problem, we first perform cross-dataset evaluations, as shown in the top three rows of Table \ref{tab:cross-dataset}.
Here, we provide two settings: 1) train the model on Matterport3D and test on Realtor360 and Stanford2D3D (the left part in Table \ref{tab:cross-dataset}), and 2) train the model on Realtor360 and test on Matterport3D and Stanford2D3D (the right part in Table \ref{tab:cross-dataset}).
Results show that our model consistently performs better than the other approaches, HorizonNet and AtlantaNet, which validate that our method is more robust to the cross-dataset setting.

Moreover, we aim to demonstrate that using the 360$^\circ$ depth datasets for pre-training is able to improve the generalizability of our proposed network. However, obtaining depth ground truth from laser scanners is much more expensive than labeling the layout ground truth, and hence we focus on adopting the synthetic Structured3D~\cite{structure3D} dataset to perform our model pre-training, in which this dataset is collected from a virtual environment and the ground truths are in high-quality and are easy to obtain.
In total, Structured3D contains 21,835 room scenes and 196,515 photo-realistic panoramas along with the corresponding ground truth depth maps, where we can extract the {\HD} maps to pre-train our model.

In the last row of Table \ref{tab:cross-dataset}, we show the results of finetuning on the training dataset and testing on the cross-dataset setting with such a pre-training scheme.
We find that this strategy significantly improves some of the settings, e.g., training on Matterport3D and testing on Realtor360 and Stanford2D3D, which demonstrates the benefit of designing a depth-based objective.
Moreover, the result in the within-dataset (``Train-Dataset'' in Table \ref{tab:cross-dataset}) setting for Realtor360 is also improved by around 1\%. However, the performance on Matterport3D (``Train-Dataset'') is slightly worse than the one without pre-training, and we argue that it is due to some specific characteristics (e.g., containing some outdoor scenes) in Matterport3D, which is less compatible with the scenes in Structured3D.
While pre-training achieves improvement in most cases, we show the potential of our {\DDR} framework that involves depth pre-training to obtain more 3D prior information, which may inspire more future research on studying the cross-dataset setting or depth pre-training.

\subsection{Effect of Ray-Casting Number}
To study the effect of the model sensitivity with respect to the number of casting rays (i.e., $M$) used in the ``Grid Re-sample'' procedure, we conduct experiments using $M=16,~64,~256,~\text{and}~ 1024$ in Table~\ref{tab:grid-resample}. From the results, while having more casting rays do provide a better approximation on the horizon-depth map (as the model performance increases from $M=64$ to $M=256$), the IoU starts to saturate when $M$ grows up to be larger than $256$. Taking the computational cost into consideration (where higher $M$ costs more), we choose to adopt $M=256$ as the default setting for all our experiments.

\section{Conclusions}
In this paper, we propose a differentiable L2D (layout-to-depth) procedure to convert the 360$^\circ$ layout representation into the 360$^\circ$ {\HD} map, thus enabling the training objective for our layout estimation network to take advantage of 3D geometric information.
We conduct extensive experiments on various datasets and achieve superior performance in comparison to several state-of-the-art baselines of monocular 360$^\circ$ layout estimation. Furthermore, as our proposed method is capable of adopting 360$^\circ$ depth datasets for model pre-training, it shows better generalizability for the cross-dataset evaluation scheme.

\vspace{-5mm}
\paragraph{Acknowledgements.} We thank iStaging for providing the Realtor360 dataset for research purpose. This project is funded by Ministry of Science and Technology of Taiwan (MOST 109-2634-F-007-016, MOST 110-2634-F-007-016, MOST 110-2636-E-009-001, MOST 110-2634-F-009-018, MOST Joint Research Center for AI Technology and All Vista Healthcare, and Taiwan Computing Cloud).

{\small
\bibliographystyle{ieee_fullname}

}

\end{document}